\documentclass[10pt,conference]{IEEEtran}

\usepackage[hidelinks]{hyperref}
\usepackage{graphicx}
\usepackage{amsmath, amssymb}
\usepackage{subcaption}
\usepackage{epstopdf}
\newcommand{\norm}[1]{\left\lVert#1\right\rVert}
\DeclareMathOperator{\sign}{sign}

\begin{document}
\title{Frank-Wolfe optimization for deep networks}

\author{
  Jakob Stigenberg
}

\maketitle

\begin{abstract}
    Deep neural networks is today one of the most popular choices in classification, regression and function approximation. However, the training of such deep networks is far from trivial as there are often millions of parameters to tune. Typically, one use some optimization method that hopefully converges towards some minimum. The most popular and successful methods are based on gradient descent. In this paper, another optimization method, Frank-Wolfe optimization, is applied to a small deep network and compared to gradient descent. Although the optimization does converge, it does so slowly and not close to the speed of gradient descent. Further, in a stochastic setting, the optimization becomes very unstable and does not seem to converge unless one uses a line search approach.
\end{abstract}
\section{Introduction}
Deep neural networks have seen a huge rise in popularity during the last decade and has seen many applications in machine learning tasks. Deep networks are essentially function approximations that map an input to an output through a number of linear and non-linear transformations. The specific transformations are determined by parameters that one seeks to alter in order to minimize the error generated by the network. The overall goal is thus to find the solution to find parameters $\vec{w}$ that solves
\begin{equation*}
    \min_{\vec{w}} f(x|\vec{w})
\end{equation*}

When dealing with machine learning tasks and function approximations, it is also important not to overfit. Through the years, a few different methods have been developed in order not to overfit. One simple, yet effective, method is to limit the set of parameters to a specific set. For example, one may introduce the constraint on the L1 norm,
\begin{equation}
    \vec{w} \in \mathcal{C}, \quad \mathcal{C} = \{\vec{x} : \norm{\vec{x}}_1 < \lambda\} \label{eq:l1constraint}
\end{equation}
and the goal is then to solve
\begin{equation*}
    \min_{\vec{w}\in\mathcal{C}} f(x|\vec{w})
\end{equation*}

Arguably, the most famous optimization methods, and perhaps most intuitive, are based on (stochastic) gradient descent, i.e. in order to find a minimum of a function $f(x)$, one iterates
\begin{equation*}
    x_{t+1} = x_t - \eta\nabla f(x_t)
\end{equation*}
where $\eta > 0$ is some constant, referred to as the learning rate. In the case of constraint optimization, the iterates are simply projected onto the set $\mathcal{C}$. However, the field of optimization has through the years proposed many different methods. In this paper, the viability of a constrained optimization method proposed in 1956 by Frank and Wolfe is discussed when applied to a deep network.
\section{The Frank-Wolfe algorithm}
The Frank-Wolfe algorithm is an optimization method for convex functions restricted to convex sets. Consider a convex function $f: \mathbb{R}^n \to \mathbb{R}$ and a convex subset $\mathcal{C} \subset \mathbb{R}^n$. To solve
\begin{equation*}
    \min_{\vec{x}\in\mathcal{C}} f(\vec{x})
\end{equation*}
the Frank-Wolfe algorithm proceeds as follows,
\begin{enumerate}
    \item Initialize $\vec{x}_0 \in \mathcal{C}$.
    \item Given $\vec{x}_t$, compute $\nabla f(\vec{x}_t)$. \label{enum:algorithmStart}
    \item Let $\vec{s}_t = \arg\min_{\vec{s}\in\mathcal{C}} \nabla f(\vec{x}_t)^T\vec{s}$ \label{enum:LMO}
    \item Update $\vec{x}_{t+1} = (1-\gamma) \vec{x}_t + \gamma \vec{s}_t$ for some $\gamma \in [0, 1]$ \label{enum:updateStep}
    \item Go to \ref{enum:algorithmStart})
\end{enumerate}
There are a number of variants of the Frank-Wolfe algorithm, see \cite{NIPS2015_5925}, however in this work only the basic algorithm in considered.
\section{Application of Frank-Wolfe algorithm on a convex function on the L1 ball} \label{sec:application_l1_convex}
In order to make the algorithm feasible, the minimization in step \ref{enum:LMO}) must be computable. For linear constraints, the minimization will always occur at a corner, since the function to be minimized is itself linear in $\vec{s}$. Hence, one only needs to evaluate the function at each corner in order to find the minimizer. Observe, that the L1 norm, presented in Eq. \ref{eq:l1constraint}, is indeed a linear constraint. For example, in $\mathbb{R}^2$ the L1-ball consists of all points $(x_1,x_2)\in\mathbb{R}^2$ restricted by the following four lines,
\begin{multline*}
    \mathcal{C} = \{(x_1, x_2) : |x_1| + |x_2| < \lambda\} \\ = \{(x_1, x_2) : |x_1+x_2| < \lambda, |x_1-x_2| < \lambda\}
\end{multline*}
Therefore, in order to compute step \ref{enum:LMO}) of the algorithm, it is sufficient to compute
\begin{equation*}
    \vec{s}_t = \arg\min_{\vec{s}\in\mathcal{D}} \nabla f(\vec{x}_t)^T \vec{s}
\end{equation*}
where
\begin{equation*}
    \mathcal{D} = \{(\pm \lambda, 0), (0, \pm \lambda)\}
\end{equation*}
In fact, given the that the corners of the L1-ball are the unit vectors, $\vec{s}$ is given by
\begin{equation*}
    [\vec{s}]_i = -\lambda\delta_{i,j}\sign\big(\big[\nabla f(\vec{x}_t)\big]_i\big) 
\end{equation*}
where
\begin{equation*}
    j = \arg\max_i |[\nabla f(\vec{x}_t)]_i|
\end{equation*}
and $[\cdot]_i$ is the $i$:th component of the vector. Therefore, in $\mathbb{R}^n$, it is possible to solve the minimization problem in only $n$ steps.

\subsection{Choice of learning parameter $\gamma$}
The parameter $\gamma$ governs the size of each step towards $\vec{s}$. Consider step \ref{enum:updateStep}) of the algorithm, with $\gamma=1$ the algorithm will show very oscillatory updates, while $\gamma=0$ yields no change (learning). From theory, there exists a few $\gamma$ for which the convergence rate of the algorithm is known to be $\mathcal{O}(T^{-1})$ \cite{jaggi2013revisiting}. Two are mentioned here,
\begin{enumerate}
    \item $\gamma_t = \frac{2}{2+t}$ \label{enum:stepsize_nonlinesearch}
    \item $\gamma_t = \arg\min_{\gamma\in[0,1]} f((1-\gamma)\vec{x}_t + \gamma \vec{s}_t)$ \label{enum:stepsize_linesearch}
\end{enumerate}
Note that the second parameter is essentially performing a line search over the 'true' function to find the optimal $\gamma$. This could easily be accomplished using projected gradient descent, since the projection is easily calculated. The derivative with respect to $\gamma$ is given by $\nabla f\big((1-\gamma)\vec{x}_t+\gamma\vec{s}_t\big)^T(\vec{s}_t-\vec{x}_t)$. Moreover, two additional step sizes are considered, one fixed step-size, $\gamma_t=C$, and one proportional to the gradient, $\gamma_t = C\norm{\nabla f(\vec{x}_t)}$. Note, however, that these two step-sizes are parameter dependent while the first two are not.

\begin{figure}
    \centering
    \includegraphics[width=\linewidth]{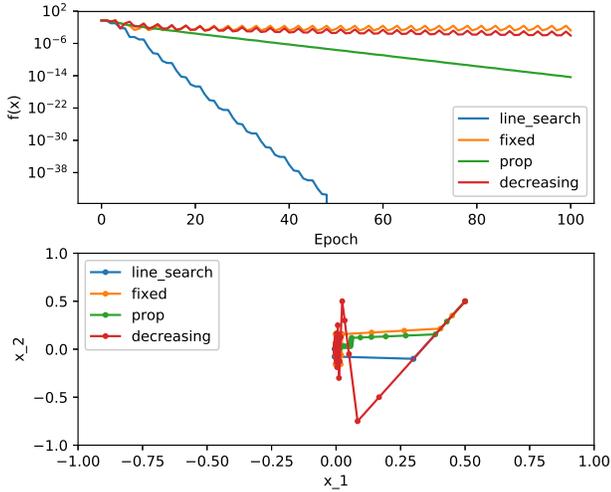}
    \caption{The function $f(x_1,x_2) = x_1^2 + x_2^2$ minimized using Frank-Wolfe optimization starting at $(x_1,x_2) = (0.5,0.5)$ using the following step sizes $\gamma_t = 0.1$ (fixed), $\gamma_t = 0.1\norm{\nabla f(\vec{x}_t)}$ (prop), $\gamma_t = 2/(2+t)$ (decreasing), $\gamma_t = \min_{\gamma\in[0,1]}f((1-\gamma)\vec{x}_t+\gamma\vec{s}_t)$ (line search).}
    \label{fig:non_linesearch}
\end{figure}

In Fig. \ref{fig:non_linesearch}, the convex function $f(x_1,x_2) = x_1^2 + x_2^2$ is optimized using the four step sizes introduced. First notice that even though the theoretical convergence results of the decreasing step size and the line search are equal, it is obvious that the line search performs much better in practice. It is interesting also to notice that a fixed step size seems to perform similar to the decreasing step size. Although a fixed step size will oscillate around a minimum, it does seem to provide a smoother path towards the vicinity of the minimum. Further, the step size proportional to the size of the gradient provides smooth path and better convergence than the decreasing one, however, it seems to be very sensitive to the parameter chosen.
\section{Non-stochastic application to a deep network}
The loss functions that one attempts to minimize when constructing a deep network are in general not convex in the parameters of the network, so the theoretical convergence results will not hold. Still, hopefully the minimization procedure might converge into local convex areas.

As an example application, consider classifying whether a point uniformly generated in the area $[0,1]^2$ is located inside or outside of the circle with radius $1$ centered around the origin. The network to optimize is a simple fully connected network with three hidden layers, each with 25 neurons. The activations are all ReLU, except for the final output which is the hyperbolic tan function. The loss function is the MSELoss. All training and test sets contain 1000 data points. Finally, the total weight vector was restricted to the L1-ball of radius 10, i.e.
\begin{equation*}
    \sum_i |w_i| < 10
\end{equation*}
where $i$ loops over every weight in every layer. Hence, the optimization problem is similar to that of Section \ref{sec:application_l1_convex}, apart from the function being non-convex. The algorithm will be compared to gradient descent, which is the goto optimization method of today. However, gradient descent runs over unconstrained problems, therefore the constraint is introduced into the loss function as a penalization term. So, when running gradient descent, the objective is
\begin{equation*}
    \min_{\vec{w}} \Big[\sum_n (f(\vec{x}_n|\vec{w}) - \vec{y}) + \frac{1}{10}\sum_i |w_i|\Big]
\end{equation*}

\subsection{Results}
\begin{figure}
    \centering
    \includegraphics[width=\linewidth]{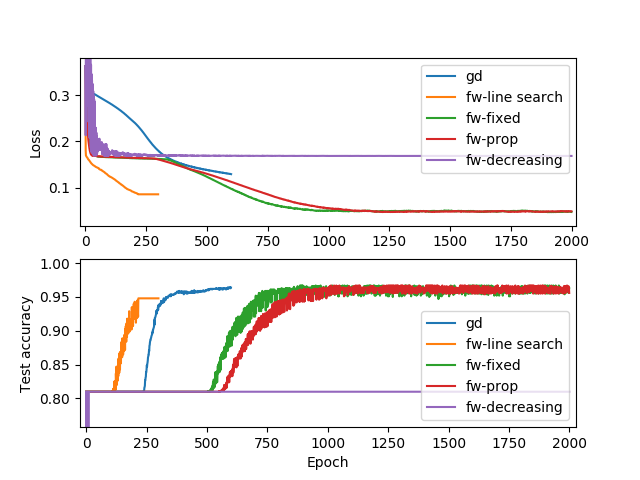}
    \caption{Full gradient descent and Frank-Wolfe optimization using different step sizes applied to a deep network.}
    \label{fig:deep_full}
\end{figure}

In Fig. \ref{fig:deep_full}, the different step sizes are applied to the deep network as well as full gradient descent with learning rate $10^{-1}$. The line search was implemented in the following way:
\begin{enumerate}
    \item Find $\lambda\in\{0,0.01,\dots,0.99\}$ minimizing $f((1-\lambda)\vec{x}+\lambda \vec{s})$.
    \item Run 100 steps of (projected) gradient descent with step size $0.01$ starting from $\gamma=\lambda$.
\end{enumerate}
$\lambda = 1$ was omitted since it is then possible to implement the algorithm without using a copy of all parameters. The fixed step-size was set to $3\cdot 10^{-3}$ and the step-size proportional to the gradient magnitude had a proportionality factor of $3\cdot 10^{-2}$.

First, notice that all but the decreasing step-size work. Through experimenting by initializing the decreasing step-size at different magnitudes, it was concluded that the decrease is too quick, i.e. the effective time during which the updates of the weights have an effect on the result is too short.

Secondly, line search seems to perform very well and reaches convergence quicker than gradient descent. However, the time required for each epoch is multiple magnitudes greater compared to gradient descent, thus the actual training time required by gradient descent is still shorter than line search.

Finally, the constant step-size and the proportional step-size behave very similarly. They both converge to approximately 95\% test accuracy, as do line search and gradient descent. However, they require significantly more epochs to reach convergence, and the time required per epoch is of the same magnitude as gradient descent. Although their convergence rate can be improved by increasing their step-sizes, it comes at the cost of a very noisy convergence. Furthermore, even if one could argue that one could have a larger step-size early on and then decrease it as convergence occurs, the same can be done to gradient descent. Therefore, even though they do work, they do not out-favor gradient descent.
\section{Stochastic application to a deep network}
Although it is nice to see that the Frank-Wolfe algorithm can perform well, the full power of gradient descent comes through its stochastic application, since it provides a good way of training using a lot of data. Therefore, if Frank-Wolfe is to have any sort of future within deep networks, it needs to function using stochastic data. For this section, the decreasing step size is omitted, since it did not work on non-stochastic data. The the same dataset is considered as in the previous section. This time, the training is done using mini-batches, instead of presenting the full dataset in each epoch.

\subsection{Results}
\begin{figure}
    \centering
    \begin{subfigure}{\linewidth}
        \includegraphics[width=\linewidth]{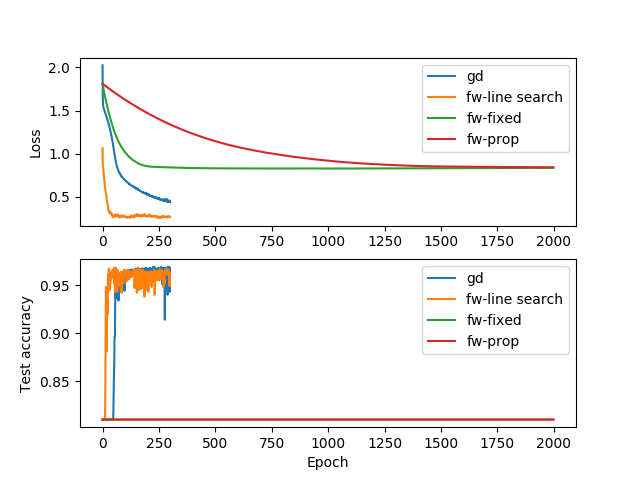}
        \caption{Batch size 200}
        \label{fig:stochastic200}
    \end{subfigure}
    \begin{subfigure}{\linewidth}
        \includegraphics[width=\linewidth]{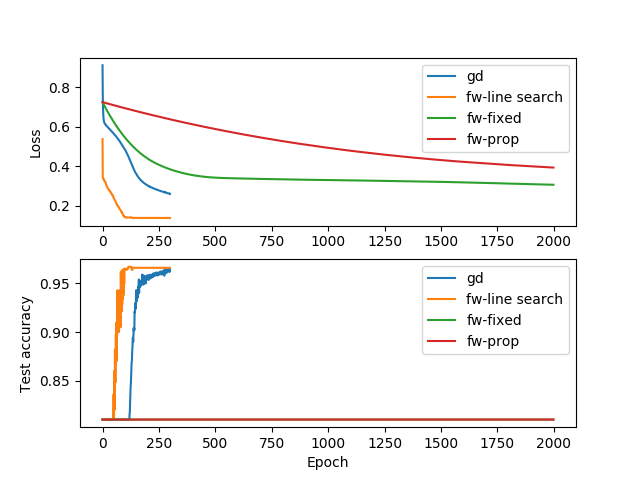}
        \caption{Batch size 500}
        \label{fig:stochastic500}
    \end{subfigure}
    \caption{Optimization done using stochastic gradient descent and stochastic Frank-Wolfe using different batch sizes}
    \label{fig:stochastic}
\end{figure}

At first, a batch size of $200$ was used. The results are shown in Fig. \ref{fig:stochastic200}. As expected, gradient descent works just fine, in this case with a learning rate of $10^{-1}$. The line search still works fine, but none of the other two step sizes seem to function using the stochastic data. A number of different constants were tried, in this plot both were set to $10^{-4}$. It seems as if the algorithm does not cope with the variance in the gradient. Increasing the batch size to $500$ as in Fig. \ref{fig:stochastic500} did not provide better results. Further, reducing the batch size to $100$, the line search stopped working as well.
\section{Conclusion}
Using the very simple dataset considered in this work, the Frank-Wolfe algorithm does work fine using a non-stochastic approach. However, doing so in large scale applications is not feasible it is therefore necessary to use stochastic methods. The algorithm seems to have a very hard time coping with the variance in the gradients. Using a line search approach seems to be the most viable option, however, it too did fail when the variance was increased (batch-size reduced). Compared to stochastic gradient descent, which works well even with batch-size one, the line search approach is \textit{much} slower. Since it requires multiple forward and backward passes done per iteration, the algorithm scales very poorly with larger models and renders it \textit{completely} un-viable when using even medium sized convolutional networks, as pointed out by \cite{reddi2016stochastic}, e.g. for the MNIST dataset. In the end, a naive stochastic implementation of the Frank-Wolfe algorithm does not yield sufficient results and stochastic gradient descent seems to be the more viable option. Modifications seem plausible, see e.g. \cite{reddi2016stochastic}.

\newpage
\bibliographystyle{IEEEtran}
\bibliography{literature}

\end{document}